\pdfoutput=1

\def\year{2022}\relax
\documentclass[letterpaper]{article} 
\usepackage{aaai22}  
\usepackage{times}  
\usepackage{helvet}  
\usepackage{courier}  
\usepackage[hyphens]{url}  
\usepackage{graphicx} 
\usepackage{natbib}  
\usepackage{caption} 
\DeclareCaptionStyle{ruled}{labelfont=normalfont,labelsep=colon,strut=off} 
\usepackage{algorithm}
\usepackage{algorithmic}
\usepackage{newfloat}
\usepackage{listings}
\floatstyle{ruled}
\newfloat{listing}{tb}{lst}{}
\floatname{listing}{Listing}
\title{Learning Sketches for Decomposing Planning Problems\\into Subproblems of Bounded Width: Extended Version}
\author{
    Dominik Drexler\textsuperscript{\rm 1},
    Jendrik Seipp\textsuperscript{\rm 1},
    Hector Geffner\textsuperscript{\rm 3,2,1}
}
\affiliations{
    \textsuperscript{\rm 1}Link{\"o}ping University, Link{\"o}ping, Sweden\\
    \textsuperscript{\rm 2}Universitat Pompeu Fabra, Barcelona, Spain\\
    \textsuperscript{\rm 3}Instituci{\'o} Catalana de Recerca i Estudis Avan\c{c}ats (ICREA), Barcelona, Spain\\
    \{dominik.drexler, jendrik.seipp\}@liu.se, hector.geffner@upf.edu
}

\usepackage{thmtools, thm-restate}  
\newtheorem{definition}{Definition}
\newtheorem{theorem}[definition]{Theorem}

\newcommand{\denselist}{\itemsep 0pt\partopsep 0pt}

\newcommand{\Rphi}{\ensuremath{R_\Phi}}

\newcommand{\inlinecite}[1]{\citet{#1}}

\newcommand{\tup}[1]{(#1)}

\newcommand{\ws}{\ensuremath{\mathit{ws}}}
\newcommand{\iw}[1]{\ensuremath{\text{IW}(#1)}\xspace}

\newcommand{\siw}{\ensuremath{\text{SIW}}\xspace}
\newcommand{\siwR}{\ensuremath{\text{SIW}_{\text R}}\xspace}
\newcommand{\siwRk}{\ensuremath{\text{SIW}_{\text R}}($k$)\xspace}

\newcommand{\lama}{\ensuremath{\text{LAMA}}\xspace}
\newcommand{\bfws}{\ensuremath{\text{BFWS}}\xspace}

\newcommand{\Q}{\mathcal{Q}}

\newcommand{\pplus}{\hspace{-.05em}\raisebox{.15ex}{\footnotesize$\uparrow$}}
\newcommand{\mminus}{\hspace{-.05em}\raisebox{.15ex}{\footnotesize$\downarrow$}}

\newcommand{\EQ}[1]{#1{\,=\,}0}
\newcommand{\GT}[1]{#1{\,>\,}0}

\newcommand{\DEC}[1]{#1\mminus}
\newcommand{\INC}[1]{#1\pplus}
\newcommand{\UNK}[1]{#1?}

\newcommand{\Omit}[1]{}

\newcommand{\prule}[2]{\{ #1 \} \mapsto \{ #2 \}}

\newcommand{\domain}{\ensuremath{D}}
\newcommand{\instance}{\ensuremath{I}}
\newcommand{\goal}{\ensuremath{\mathit{Goal}}}
\newcommand{\initial}{\ensuremath{\mathit{Init}}}

\newcommand{\states}{\ensuremath{S}}
\newcommand{\initialstate}{\ensuremath{s_0}}
\newcommand{\goalstates}{\ensuremath{G}}
\newcommand{\actions}{\ensuremath{\mathit{Act}}}
\newcommand{\successor}{\ensuremath{f}}
\newcommand{\applicability}{\ensuremath{A}}

\newcommand{\Ps}{{\cal P}}
\newcommand{\sattheory}{\ensuremath{T_{k,m}({\cal P},{\cal F})}}
\newcommand{\satselect}[1]{\ensuremath{\mathit{select(#1)}}}
\newcommand{\satcond}[3]{\ensuremath{\mathit{cond}(#1,#2,#3)}}
\newcommand{\sateff}[3]{\ensuremath{\mathit{eff}(#1,#2,#3)}}
\newcommand{\satsubgoal}[2]{\ensuremath{\mathit{subgoal(#1,#2)}}}
\newcommand{\satsubgoals}[3]{\ensuremath{\mathit{subgoals(#1,#2,#3)}}}
\newcommand{\satrule}[3]{\ensuremath{\mathit{sat}\_\mathit{rule}(#1,#2,#3)}}

\newcommand{\vatrobot}[1]{\ensuremath{\mathit{at}\text{-}\mathit{robot}(#1)}}

\newcommand{\sat}[2]{\ensuremath{\mathit{at}(#1,#2)}}
\newcommand{\scarrying}[1]{\ensuremath{\mathit{carrying}(#1)}}

\newcommand{\stightened}[1]{\ensuremath{\mathit{tightened}(#1)}}

\newcommand{\gatrobot}[1]{\ensuremath{\mathit{at}\text{-}\mathit{robot}(#1)}}
\newcommand{\gat}[2]{\ensuremath{\mathit{at}(#1, #2)}}
\newcommand{\gcarrying}[1]{\ensuremath{\mathit{carrying}(#1)}}

\newcommand{\bholding}[1]{\ensuremath{\mathit{hold}(#1)}}
\newcommand{\bontable}[1]{\ensuremath{\mathit{on}\text{-}\mathit{table}(#1)}}
\newcommand{\bon}[2]{\ensuremath{\mathit{on}(#1, #2)}}
\newcommand{\bclear}[1]{\ensuremath{\mathit{clear}(#1)}}

\newcommand{\cglutenfreesandwich}[1]{\ensuremath{\mathit{gluten}\text{-}\mathit{free}\text{-}\mathit{sandwich}(#1)}}
\newcommand{\catkitchensandwich}[1]{\ensuremath{\mathit{at}\text{-}\mathit{kitchen}\text{-}\mathit{sandwich}(#1)}}
\newcommand{\cnotexist}[1]{\ensuremath{\mathit{notexist}(#1)}}
\newcommand{\cat}[2]{\ensuremath{\mathit{at}(#1, #2)}}
\newcommand{\contray}[2]{\ensuremath{\mathit{ontray}(#1, #2)}}
\newcommand{\cserved}[1]{\ensuremath{\mathit{served}(#1)}}
\newcommand{\ckitchen}{\ensuremath{\mathit{kitchen}}}

\newcommand{\mliftat}[1]{\ensuremath{\mathit{lift}\text{-}\mathit{at}(#1)}}
\newcommand{\mboarded}[1]{\ensuremath{\mathit{boarded}(#1)}}
\newcommand{\mserved}[1]{\ensuremath{\mathit{served}(#1)}}

\usepackage{booktabs}
\usepackage{enumerate}
\usepackage{xspace}
\usepackage{todonotes}
\usepackage{amssymb}
\usepackage{amsmath}
\usepackage[title]{appendix}
\presetkeys{todonotes}{inline}{}

\begin{document}

\maketitle

\begin{abstract}
Recently, sketches have been introduced as a general language for representing the subgoal structure of instances
drawn from the same domain. Sketches are collections of rules of the form $C\mapsto E$ over a given set of features where $C$ expresses Boolean
conditions and $E$ expresses qualitative changes. Each sketch rule defines a
subproblem: going from a state that satisfies $C$ to a state that
achieves the change expressed by $E$ or a goal state. Sketches can encode
simple goal serializations, general policies, or decompositions of bounded width that can be solved greedily, in
polynomial time, by the \siwR variant of the \siw algorithm.
Previous work has shown the computational value of sketches over  benchmark domains that,
while tractable, are challenging for domain-independent planners.
In this work, we address the problem of
learning sketches automatically given a planning domain, some instances
of the target class of problems, and the desired bound on the sketch
width. We present a logical formulation of the problem, an
implementation using the ASP solver Clingo, and experimental results.
The sketch learner and the \siwR planner yield a domain-independent planner
that learns and exploits domain structure in a crisp and explicit form.
\end{abstract}

\section{Introduction}

Classical planners manage to solve problems that span exponentially
large state spaces by exploiting problem structure.
Domain-independent methods make implicit assumptions about structure, such as subgoals being independent (delete-relaxation)
or having low width (width-based search).
Domain-dependent methods, on the other hand, usually make problem structure explicit in the form of hierarchies
that express how tasks decompose into subtasks \cite{erol-et-al-aaai1994,georgievski-aiello-aij2015,bercher-et-al-ijcai2019}.

An alternative, simpler language for representing problem structure explicitly has been introduced recently in the form
of \emph{sketches} \cite{bonet-geffner-aaai2021}. Sketches are collections of rules of the form $C \mapsto E$ defined over a
given set of Boolean and numerical domain features $\Phi$ where $C$ expresses Boolean conditions on the features, and $E$ expresses
qualitative changes in their values.
Each sketch rule expresses a subproblem: the problem of going from a state $s$ whose feature values satisfy the condition $C$,
to a state $s'$ where the feature values change in agreement with $E$.

The language of sketches is powerful, as sketches can encode everything from
simple goal serializations to full general policies.
Indeed, the language of general policies is the language of sketches
but with a slightly different semantics where the subgoal states $s'$ to be reached from a state $s$ are
restricted to be one step away from $s$ \cite{bonet-geffner-ijcai2018,frances-et-al-aaai2021}.
More interestingly, sketches can split problems into subproblems of \emph{bounded width}
\cite{lipovetzky-geffner-ecai2012,lipovetzky-ijcai2021} which can then be solved greedily,
in polynomial time, by a variant of the SIW algorithm,
called \siwR \cite{bonet-geffner-aaai2021}.
The computational value of sketches crafted by hand
has been shown over several planning domains which, while tractable,
are challenging for domain-independent planners \cite{drexler-et-al-kr2021}.

In this work, we build on these threads (general policies, sketches, and width) to
address the problem of \emph{learning sketches automatically}. For this, the inputs
are the planning domain, some instances, and the desired bound $k$ on width,
usually $k=0,1,2$. The width $k$ of the sketch relative to a class of problems
bounds the width of the resulting subproblems that can be solved greedily in time and space
that are exponential in $k$.
%
In order to address the problem of learning sketches, we present a logical formulation,
an implementation of the learner on top of the ASP solver Clingo \cite{gebser-et-al-tplp2019}, and experimental results.
We start  with an example and a review of planning, width, and sketches.

\section{Example}

For an illustration of the concepts and results, before presenting the formal definitions,
let us recall a simple domain called Delivery \cite{bonet-geffner-aaai2021},
where an agent moves in a grid to pick up packages and deliver them to a target cell,
one by one. A general policy $\pi$ for this class of instances can be expressed in terms
of the set of features $\Phi=\{H,p,t,n\}$ that express ``holding a package'', ``distance to the nearest package'',
``distance to the target cell'', and ``number of undelivered packages'', respectively.
A domain-independent method for generating a large pool of features from the domain predicates
that include these four is given by \inlinecite{bonet-et-al-aaai2019}. Provided with the features in $\Phi$,
a \textbf{general policy} for the whole class $\Q_D$ of Delivery problems (any grid size, any number of packages in
any location, and any location of the agent or the target) is given by the rules:

\vspace{0.2cm}
\begin{center}
\begin{tabular}{ll}
$\prule{\neg H,\GT{p}}{\DEC{p},\UNK{t}}$ & ; go to nearest pkg \\
$\prule{\neg H, \EQ{p}}{H}$ & ; pick it up \\
$\prule{H,\GT{t}}{\DEC{t}}$ & ; go to target \\
$\prule{H,\GT{n},\EQ{t}}{\UNK{H}, \DEC{n}, \UNK{p}}$ & ; deliver pkg\\
\end{tabular}
\end{center}
\vspace{0.2cm}
The rules say to perform  any action that  decreases  the distance to the nearest package ($\DEC{p}$), no matter the effect
on the distance to target ($\UNK{t}$), when not holding a package ($\neg H$);
to pick a package when possible, making $H$ true;
to go to the target cell when holding a package, decrementing $t$;
and to drop the package at the target, decrementing $n$, making $H$ false, and affecting $p$.
Expressions $p?$ and $m?$ mean that $p$ and $m$ can change in any way,
while no mention of a feature in the effect of a rule means that
the value of the feature must not change.

One can show that the general policy above solves any instance of Delivery,
or alternatively, that it represents a  \textbf{width-zero sketch}, where the subproblem
of going from a non-goal state $s$ to a state $s'$ satisfying a rule $C \mapsto E$
can always be done in one step, leading greedily to the goal $(n=0)$.
A  pair of states $(s,s')$ satisfies  a rule $C \mapsto E$ when the feature
values in $s$ satisfy the conditions in $C$, and the change in feature values
from $s$ to $s'$ is compatible with the changes and no-changes expressed in $E$.
A \textbf{width-2 sketch}   can be defined instead by means of a single rule
and a single feature $\Phi=\{n\}$:

\vspace{0.2cm}
\begin{center}
\begin{tabular}{ll}
$\prule{\GT{n}}{\DEC{n}}$ \ \ \ \ \ \ & ; deliver pkg\\
\end{tabular}
\end{center}
\vspace{0.2cm}
The subproblem of going from a state $s$ where $n(s) > 0$ holds,
to a state $s'$ for which $n(s') < n(s)$ holds, has width bounded by $2$, as these subproblems,
in the worst case, involve moving to the location of a package, picking it up, moving to the target, and dropping it.
Halfway between the width-0 sketch represented by general policies, and the width-2 sketch above,
is the  \textbf{width-1 sketch}  defined with two rules and two  features, $\Phi=\{H,n\}$ as:

\vspace{0.2cm}
\begin{center}
\begin{tabular}{ll}
$\prule{\neg H}{H}$ & ; pick pkg \\
$\prule{H,\GT{n}}{\UNK{H},\DEC{n}}$ & ; deliver pkg\\
\end{tabular}
\end{center}
\vspace{0.2cm}
The first rule captures the subproblem of getting hold of a package, which may involve
moving to the package and picking it up, with width 1; the second rule captures
the subproblem of delivering the package being held, which may involve moving to the target
and dropping it there, also with width 1. Since every non-goal state is covered by
one rule or the other, the width of the sketch is 1, which means that any instance of Delivery
can be solved greedily by solving subproblems of width no greater than 1 in linear time.

\Omit{
Domains with width $k$-sketches are solved in time that is exponential in $k$, and not the number
of problem variables, by the \siwR algorithm, which will iteratively run the IW($w$) algorithm
for values of $w$ that will not exceed $k$, always reach the problem goal.
}

These are all handcrafted sketches. The  methods to be formulated below will learn similar sketches
when given the planning domain, a few instances, and the desired bound on sketch width.



\section{Background}

We review classical planning, width, and sketches  drawing from \inlinecite{lipovetzky-geffner-ecai2012},
\inlinecite{bonet-geffner-aaai2021}, and
\inlinecite{drexler-et-al-kr2021}.

\subsection{Classical Planning}

A \emph{planning problem} or \emph{instance} is a pair $P=\tup{\domain,\instance}$
where $\domain$ is a first-order \emph{domain} with action schemas defined over predicates,
and $\instance$ contains  the  objects in the instance and two sets of ground literals,
the initial and goal situations  $\initial$ and  $\goal$.
The initial situation is consistent and complete, meaning either a ground literal or its complement
is in $\initial$.
An instance $P$ defines a state model $S(P)=\tup{\states,\initialstate,\goalstates,\actions,\applicability,\successor}$ where
the states in $S$ are the truth valuations over the ground atoms represented by the set of literals that they make true,
the initial state $s_0$ is $\initial$, the set of goal states $\goalstates$ are those that make the goal literals in $\goal$ true,
and the actions $Act$ are the ground actions obtained from the schemas and objects. The ground actions in
$\applicability(s)$ are the ones that are applicable in a state $s$; namely, those whose preconditions are true in $s$,
and the state transition function $f$ maps a state $s$ and an action $a \in \applicability(s)$ into the successor state $s'=f(a,s)$.
A \emph{plan} $\pi$ for $P$ is a sequence of actions $a_0,\ldots,a_n$ that is executable in $s_0$ and maps the initial state $s_0$
into a goal state; i.e., $a_i \in \applicability(s_i)$, $s_{i+1}=f(a_i,s_i)$, and $s_{n+1}\in\goalstates$.
A state $s$ is \emph{solvable} if there exists a plan starting at $s$, otherwise it is \emph{unsolvable} (also called \emph{dead-end}).
Furthermore, a state $s$ is \emph{alive} if it is solvable and it is not a goal state.
The \emph{length} of a plan is the number of its actions, and a plan is \emph{optimal} if there is no shorter plan.
Our objective is to find suboptimal plans for \emph{collections} of instances $P=\tup{\domain,\instance}$
over fixed domains $\domain$ denoted as $\Q_\domain$ or simply as $\Q$.

\subsection{Width}

The simplest width-based search method for solving a planning problem $P$ is \iw{1}.
It is a standard breadth-first search in the rooted directed graph associated with the state model
$S(P)$ with one modification: \iw{1} prunes a newly generated state if it does not make
an atom true for the first time in the search. The procedure \iw{k} for $k > 1$ is like \iw{1}
but prunes a state if a newly generated state does not make a collection of up to $k$ atoms true for the first time.
Underlying the IW algorithms is the notion of \emph{problem width} \cite{lipovetzky-geffner-ecai2012}:
\begin{definition}[Width]
 \label{def:width}
 The \textbf{width} $w(P)$ of a classical planning problem $P$ is the minimum $k$ for which there exists a
 \textbf{sequence $t_0,t_1,\ldots,t_m$ of atom tuples $t_i$} from $P$,
 each consisting of at most $k$ atoms, such that:
 \begin{enumerate}
 \item $t_0$ is true in the initial state $\initialstate$ of $P$,
 \item any optimal plan for $t_i$ can be extended into an optimal plan for $t_{i+1}$ by adding a single action, $i=1,\ldots,n-1$,
 \item if $\pi$ is an optimal plan for $t_m$, 
 $\pi$ is an optimal plan for $P$.
 \end{enumerate}
\end{definition}

If a problem $P$ is unsolvable, $w(P)$ is set to the number of variables
in $P$, and if $P$ is solvable in at most one step, $w(P)$ is set to $0$
\cite{bonet-geffner-aaai2021}. Chains of tuples $\theta = (t_0, t_1, \dots, t_m)$
that comply with conditions 1--2 are called \textbf{admissible}, and the
size of $\theta$ is the size $|t_i|$ of the largest tuple in the chain.
The width $w(P)$ is thus the minimum size of an admissible chain for $P$
that is also optimal (condition 3). Furthermore, the width of a conjunction of atoms $T$ (or arbitrary set of states $S'$ in $P$)
is the width of a problem $P'$ that is like $P$ but with the goal $T$ (resp. $S'$).\footnote{In the literature, a chain is admissible
  when it complies with conditions 1--3. The reason for dropping condition 3 will become clear when we introduce
  the notion of ``satisficing width''.}


The \iw{k} algorithm expands up to $N^k$ nodes, generates up to $bN^k$ nodes, and runs in time and space
$O(bN^{2k-1})$ and $O(bN^k)$, respectively, where $N$ is the number of atoms and $b$ is a bound on the branching factor of the problem $P$.
\iw{k} is guaranteed to solve $P$ optimally if $w(P) \leq k$.
If the width of $P$ is not known, the \textbf{IW} algorithm can be run instead which calls \iw{k} iteratively for
$k\!=\!0,1,\ldots,N$ until the problem is solved, or found to be unsolvable.

For problems with conjunctive goals, the \textbf{SIW} algorithm \cite{lipovetzky-geffner-ecai2012}
starts at the initial state $s=s_0$ of $P$, and performs an IW search
from $s$ to find a shortest path to a state $s'$ such that
$\#g(s') < \#g(s)$ where $\#g(s)$ counts the number of unsatisfied top-level goals of $P$
in state $s$. If $s'$ is not a goal state, $s$ is set to $s'$
and the loop repeats.

\Omit{
In practice, the \iw{k} searches in SIW are limited to $k \leq 2$ or $k \leq 3$, so that SIW solves
a problem or fails in low polynomial time. SIW performs well in many benchmark domains but
fails in problems where the width of some top goal is not small, or the top
goals can't be serialized greedily. More recent methods
address these limitations by using width-based notions (novelty measures) in complete best-first search algorithms
\cite{lipovetzky-et-al-aaai2017,frances-et-al-ijcai2017}, yet they also struggle in problems where some top goals have high width.
In this work, we take a different route: we keep the greedy polynomial searches
underlying SIW but consider a richer class of problem decompositions expressed through sketches.
The resulting planner \siwR is \textbf{not} domain-independent like SIW,
but it illustrates that a bit of domain knowledge can go a long way in the effective solution of arbitrary domain instances.
}

\subsection{Sketches}


A \textbf{feature} is a function of the state over a class of problems $\Q$.
The features considered in the language of sketches are {Boolean},
taking values in the Boolean domain, or {numerical},
taking values in the non-negative integers.
For a set of features $\Phi = \{f_1,\ldots,f_N\}$
and a state $s$ of some instance $P$ in $\Q$,
we denote the \textbf{feature valuation} determined by a state
$s$ as $f(s) = (f_1(s),\ldots,f_N(s))$, and arbitrary feature valuations
as $f$ and $f'$.

\Omit{
A {Boolean} feature valuation over $\Phi$
refers instead to the valuation of the expressions $p$ and $n=0$
for Boolean and numerical features $p$ and $n$ in $\Phi$.
If $f$ is a feature valuation, $b(f)$ will denote the Boolean feature valuation
determined by $f$ where the values of numerical features are just compared with $0$.
The set of features $\Phi$ \textbf{distinguish} or \textbf{separate}
the goals in $\Q$ if there is a set $B_\Q$ of Boolean feature valuations
such that $s$ is a goal state of an instance $P \in \Q$ iff $b(f(s)) \in B_\Q$.
}

\Omit{
For example, if $\Q_{clear}$ is the set of all blocks world instances with stack/unstack operators
and common goal $clear(x) \land handempty$ for some block $x$, and $\Phi=\{n(x),H\}$
are the features that track the number of blocks above $x$ and whether
the gripper is holding a block, then there is a single Boolean goal valuation that
makes the expression $n(x)=0$ true and $H$ false.
}

A \textbf{sketch rule} over features $\Phi$ has the form $C \mapsto E$
where $C$ consists of Boolean feature conditions, and $E$ consists of feature effects.
A Boolean (feature) condition is of the form $p$ or $\neg p$ for a Boolean feature $p$ in $\Phi$,
or $n=0$ or $n>0$ for a numerical feature $n$ in $\Phi$.
A feature effect is an expression of the form $p$, $\neg p$,
or $\UNK{p}$ for a Boolean feature $p$ in $\Phi$, and $\DEC{n}$, $\INC{n}$,
or $\UNK{n}$ for a numerical feature $n$ in $\Phi$.
The syntax of sketch rules is the syntax of the policy rules
used to define generalized policies \cite{bonet-geffner-ijcai2018},
but their semantics is different.
In policy rules, the effects have to be delivered in one step by state transitions,
while in sketch rules, they can be delivered by longer state sequences.


A pair of feature valuations of two states $(f(s),f(s'))$,
referred to as $(f, f')$, \textbf{satisfies a sketch rule} $C \mapsto E$ iff
1)~$C$ is true in $f$,
2)~the Boolean effects $p$ ($\neg p$) in $E$ are true in $f'$,
3)~the numerical effects are satisfied by the pair $(f,f')$; i.e.,
if $\DEC{n}$ in $E$ (resp. $\INC{n}$), then the value of $n$ in $f'$ is smaller (resp. larger) than in $f$, and
4)~features that do not occur in $E$ have the same value in $f$ and $f'$.
Adding the effects $\UNK{p}$ and $\UNK{n}$ allows the values of features $p$ and $n$
to change in any way. In contrast, the value of features
that do not occur in $E$ must be the same in $s$ and $s'$.

A sketch is a collection of sketch rules that establishes a ``preference ordering'' `$\prec$' over feature valuations
where $f' \prec f$ if the pair of feature valuations $(f,f')$ satisfies a rule. If the sketch is \textbf{terminating},
then this preference order is a strict partial order: irreflexive and transitive.
Checking termination requires time that is exponential in the number of features \cite{bonet-geffner-aaai2021}.

Following Bonet and Geffner, we do not use these orderings explicitly but the associated problem \textbf{decompositions}.
The set of \textbf{subgoal states} $G_R(s)$ associated with a sketch $R$ in a state $s$ of a problem $P \in Q$,
is the set of states $s'$ that comprise the goal states of $P$
along with those with feature valuation $f(s')$ such that the pair $(f(s),f(s'))$ satisfies a rule in $R$.
The set of states $s'$ in $G_R(s)$ that are closest to $s$ is denoted as $G^*_R(s)$.


\subsection{Sketch Width}

The \textbf{\siwR} algorithm is a variant of  SIW that uses a given
sketch $R$ for solving problems $P$ in $\Q$. \siwR starts at the initial state $s=s_0$ of $P$
and then runs an IW search to find a state $s'$ in $G_R(s)$.
If $s'$ is not a goal state, then $s$ is set to $s'$,
and the loop repeats until a goal state is reached.
The \siwRk algorithm is like {\siwR} but calls
the procedure IW($k$) internally, not IW.
%
\Omit{
The only difference between \siw and \siwR is that in \siw\
each IW search finishes when the goal counter $\#g$ is decremented,
while in \siwR, when a goal or subgoal state is reached.
The behavior of plain SIW can be emulated in \siwR using the
single sketch rule $\prule{\GT{\#g}}{\DEC{\#g}}$ in $R$ when the goal counter $\#g$ is the only
feature, and the rule $\prule{\GT{\#g}}{\DEC{\#g},\UNK{p},\UNK{n}}$, when $p$
and $n$ are the other features. This last rule says that it is always ``good'' to decrease
the goal counter independently of the effects on other features, or alternatively,
that decreasing the goal counter is a subgoal from any state $s$ where $\#g(s)$ is positive.
}

For bounding the complexity of these algorithms, let us  assume without loss of generality that
the class of problems $\Q$ is \emph{closed} in the sense that if $P$ belongs to $\Q$ so
do the problems $P'$ that are like $P$ but with initial states that are reachable in $P$
and which are not dead-ends. Then the width of the sketch $R$ over $\Q$
can be defined as follows \cite{bonet-geffner-aaai2021}:\footnote{Our definition is simpler than those used by \inlinecite{bonet-geffner-aaai2021}, and
  \inlinecite{drexler-et-al-kr2021}, as it avoids a recursive condition of the set of subproblems $P'$.
  However, it adds an extra condition that involves the dead-end states, the states from which the goal cannot be reached.
}

\begin{definition}[Sketch width]
 \label{def:sketch_width}
  The \textbf{width of sketch} $R$ over a closed class of problems $\Q$ is $w_{R}(\Q) = \max_{P \in \Q} w(P')$
  where $P'$ is  $P$ but with goal states $G^*_R(s)$ and  $s$ is the initial state of both,
  provided that $G^*_R(s)$ does not contain dead-end states.
\end{definition}


If the sketch width is bounded, \siwRk solves the instances in $\Q$
in polynomial time:\footnote{Algorithm \siwRk is needed here instead of \siwR
because the latter does not ensure that the subproblems $P'$ are solved optimally.
This is because IW($k'$) may solve problems of width $k$ non-optimally if $k' < k$.}

\begin{theorem}
 \label{thm:sketch_width}
 If $w_{R}(\Q) \leq k$ and the sketch $R$ is terminating, then  \siwR($k$)  solves the instances in $\Q$ in
 $O(bN^{|\Phi|+2k-1})$ time and $O(bN^k+N^{|\Phi|+k})$ space, where $|\Phi|$ is the number of features,
 $N$ is the number of ground atoms, and $b$ is the branching factor.
\end{theorem}

For these bounds, the features are assumed to be linear in $N$; namely, they must have
at most a linear number of values in each instance,
all computable in linear time.



\subsection{Extensions}

We finish this review with a slight generalization of Theorem~\ref{thm:sketch_width} that is worth making explicit.
For this, let us introduce a variant of the notion of width, called  \emph{satisficing width} or \emph{s-width}.
A problem $P$ has satisficing width $\leq k$ if there is an admissible chain of tuples $\tau:t_0,\ldots,t_m$ of size no greater than $k$
such that the optimal plans for tuple $t_m$ are plans for $P$.
In such a case, we say that $\tau$ \emph{is an admissible $k$-chain} for $P$.
The difference to the (standard) notion of width is that optimal plans for $t_m$ are required to be plans for $P$ but not optimal.
The result is that IW($k$) will solve problems of s-width bounded by $k$, written $\ws(P)\leq k$, but not optimally.
A convenient property is that a problem $P$ with s-width bounded by $k$
has the same bound when the set of goal states of $P$ is extended with more states.
The length of the plans computed by IW($k$) when $k$ bounds the s-width of $P$
is bounded in turn by the length $m$ of the shortest admissible $k$-chain $t_0,\ldots,t_m$ for $P$.
If the subproblem of reaching  a state $s'$ in $G_R(s)$ has s-width $k$, and $G_R^k(s)$ stands
for the  states $s'$ in $G_R(s)$ that are  no more than $m$ steps away from $s$,
the \emph{satisficing width} of the a sketch $R$ can be defined as:

\begin{definition}[Sketch s-width]
 \label{def:sketch_swidth}
 The \textbf{s-width of sketch} $R$ over a closed class of solvable problems $\Q$ is bounded by $k$, $\ws_{R}(\Q) \leq k$,
 if $\max_{P \in \Q} \ws(P') \leq k$  where $P'$ is  $P$ but with goal states $G^k_R(s)$,  and  $s$ is the initial state of $P$,
  provided that $G^k_R(s)$ does not contain dead-end states.
\end{definition}

This definition just replaces the width of subproblems by the weaker s-width, and the
subgoal states $G_R^*(s)$ by $G_R^k(s)$.

Let us finally say that a sketch $R$ is state (resp. feature) \textbf{acyclic} in $\Q$
when there is no sequence of states $s_1, \ldots, s_n$ over a problem in $\Q$  (resp. feature valuations $f(s_0)$, \ldots, $f(s_n)$),
$s_{i+1} \in G_R(s_i)$, such that $s_n = s_j$, $j < n$ (resp. $f(s_n)=f(s_j)$, $j < n$).
\inlinecite{bonet-geffner-aaai2021} showed that termination implies feature acyclicity,
and it is direct to show that feature acyclicity implies state acyclicity. 
Theorem~\ref{thm:sketch_width} can then be rephrased as:

\begin{theorem}
 \label{thm:sketch_swidth}
 If $\ws_{R}(\Q) \leq k$ and the sketch $R$ is (state) acyclic in $\Q$, then  \siwRk solves the instances in $\Q$ in
 $O(bN^{|\Phi|+2k-1})$ time and $O(bN^k+N^{|\Phi|+k})$ space, where $|\Phi|$ is the number of features,
 $N$ is the number of ground atoms, and $b$ is the branching factor.
\end{theorem}

\Omit{
In the implementation  below, bounded s-width and acyclicity will be enforced on the learned sketches over the training instances,
and the computational value of the sketches will be exploited via the \siwR algorithm  that runs faster than \siwRk, at the potential
(theoretical) risk of finding longer plans and ending in dead-end states. From a theoretical point of view, while the
properties of bounded s-width and acyclicity are not guaranteed over the test instances, we will prove that in the domains
considered, they hold, and moreover, not just bounded s-width but standard width as well.
}

\Omit{
State and feature acyclicity can be shown in a number of ways. For example, if  effects $n_i\pplus$ (resp. $n_i\mminus$; $p$) appear in each of the
sketch rules, and the rules  contain no ``opposite'' effects $n_i\mminus$ or $n_i?$ (resp. $n\pplus$ or $n?$; $\neg p$ or $p?$), then we have feature acyclicity
and hence also state acyclicity. If not, rules containing such effects can be removed, and the procedure  applied to the rules left until no sketch
rules are left. Also, one can consider a vector $v(s)$ of features that are not necessarily part of the sketch and show that that for  any pair  of states
$s_i$ and $s_{i+1}$ such that $s_{i+1}  \in G_R(s)$, $v(s) \prec v(s')$ where ``$\prec$'' is some partial order among these vectors.

In the next sections, we address the problem of learning sketches that can be exploited computationally in the solutions of a class of problems $\Q$
where the conditions of Theorem~\ref{thm:sketch_width} or \ref{thm:sketch_swidth} hold.
}

\section{Learning Sketches: Formulation}

We turn to the problem of learning sketches given a set of instances $\cal P$ of the target class of problems $\Q$
and the desired bound $k$ on sketch width. We roughly follow the approach for learning general policies \cite{bonet-et-al-aaai2019,frances-et-al-aaai2021}
by constructing  a theory $\sattheory$  from  $\cal P$, $k$, a bound $m$ on the number of sketch rules, and
a finite pool of features $\cal F$ obtained from the domain predicates and a fixed grammar.

\Omit{
The learned sketch $R$ is then read off from the optimal truth assignments that satisfies the theory,
with the cost of assignments taking into account the complexity of the features selected (number grammar rules needed to generate them)
and the size of the sketch (number of rules).
Our encoding differs to \inlinecite{frances-et-al-aaai2021} by
additionally encoding an upper bound on the number of rules to be learned.
This has the benefit that we can get rid of the costly constraint called D2-separation.
D2-separation is a more general way to test whether the classifier of transitions (or state pairs)
can be described as a general policy (or policy sketch).
We explain the learning formulation by specifying the theory and the costs to optimize.
}

\subsection{Theory}

Symbols $s, s', s''$, $f$, $v$, $t$, and $i$  refer to reachable states, features,
Boolean values, tuples of at most $k$ atoms, and sketch rules $C_i \mapsto E_i$, respectively.
States and tuples are unique to each training instance $P_j$ in $\cal P$ and thus are not tagged
with their instance. The variables (atoms) in $\sattheory$ are

\begin{itemize} \denselist
\item $\satselect{f}$ for features $f$ in pool $\cal F$,
\item $\satcond{i}{f}{v}$: for $f$, rule $i=1, \ldots, m$, $v$ Bool or '$?$',
\item $\sateff{i}{f}{v}$: for $f$, rule $i=1, \ldots,m$,  $v$ Bool or '$?$',
\item $\satsubgoal{s}{t}$: for $t$ of {width} $\leq k$ from $s$,
\item $\satsubgoals{s}{t}{s'}$: if optimal plan from $s$ to $t$ ends in $s'$,
\item $\satrule{s}{s'}{i}$: for $s'$ reachable from $s$, $i=1, \ldots m$.
\end{itemize}
When true, these atoms represent that $f$ is a feature used in the sketch,
that the $i$-th rule $C_i \mapsto E_i$ has $f$ as a condition with value $v$, or no condition if $v=?$,
respectively an effect on $f$ with value $v$, possibly `$?$',\footnote{For a numerical feature $f$,
a condition with value $v=0$ stands for $f=0$ while an effect with the same value
stands for $\DEC{f}$. Similarly for $v=1$.} that $t$ is a subgoal selected
from $s$, possibly leading to state $s'$, and that the transition from $s$ to $s'$
(not necessarily a 1-step transition) satisfies rule $i$.
\textbf{Constraints} C1--C8 capture these meanings:

\begin{enumerate}[\bfseries\upshape C1] \denselist
\item $\satcond{i}{f}{v}$, $\sateff{i}{f}{v}$ use unique $v$, imply $\satselect{f}$
\item $\vee_{t} \satsubgoal{s}{t}$, each alive $s$ has some subgoal $t$
\item $\satsubgoal{s}{t}$ iff $\wedge_{s'} \satsubgoals{s}{t}{s'}$
\item $\satsubgoals{s}{t}{s'}$ implies $\vee_{i=1,m} \satrule{s}{s'}{i}$ 
\item $\satrule{s}{s''}{i}$ implies $\vee_{t: d(s,t) < d(s,s'')} \satsubgoal{s}{t}$ 
\item $\satrule{s}{s'}{i}$ implies   $\vee_{t: d(s,t)\leq d(s,s')} \satsubgoal{s}{t}$ 
\item $\satrule{s}{s'}{i}$ iff $(s,s')$ compatible with rule $i$
\item collection of rules $i=1, \ldots, m$ is terminating
\end{enumerate}

In the constraints above, $s$ is always an alive state, $t$ is a tuple in the tuple graph rooted at $s$,
$s'$ is a state that results from an optimal plan for $t$ from $s$, and $s''$ is a dead-end state.
Checking the constraints involves constructing a tuple graph \cite{lipovetzky-geffner-ecai2012}
and labeling dead-ends in the training instances in a preprocessing phase that is exponential in $k$.
The interpretation of the constraints is direct. The first constraint says that a feature is selected if it is used in some rule.
Constraints 2--4 say that every (non-goal) solvable state $s$ ``looks'' at some subgoal $t$ of width no greater than $k$ from $s$, and
that if $s'$ may result from an optimal plan from $s$ to $t$, then the transition $(s,s')$ must satisfy one of the $m$ rules.
Constraint 5 says that if  a transition from $s$ to a dead-end state $s''$ satisfies a rule, then there must be a selected
subgoal $t$ of $s$ that is closer from $s$ than $s''$ ($d(s,t)$ and $d(s,s'')$ encode these distances, known after preprocessing).
Constraint 6 says that if a pair of states $(s,s')$ satisfies a rule, then $s'$ is a state that results from an optimal plan
from $s$ to a subgoal $t$ of $s$, or it is further from $s$ than any such state.
Constraint 7, not fully spelled out, captures the conditions
under which a pair of states $(s,s')$ satisfies sketch rule $i$ given
by the atoms $\satcond{i}{f}{v}$ and $\sateff{i}{f}{v}$.
Last, Constraint 8, not fully spelled out either,
demands that the sketch defined by these atoms is \emph{structurally terminating} \cite{bonet-geffner-aaai2021}.
Encoding structural termination requires checking acyclicity in the policy graph
that has size exponential in $|\cal F|$.
The resulting theory is sound and complete in the following sense:

\begin{restatable}[Soundness and Completeness]{theorem}{theoremtheory}
    \label{theorem:sound-complete}
    A terminating sketch $R$ with rules $C_i \mapsto E_i$, $i=1,\ldots, m$, over features $F \in {\cal F}$,
    has sketch width $w_R(\Ps^*)\leq k$ over the closed class of problems $\Ps^*$ iff the theory $\sattheory$ is satisfiable
    and has a model where the rules that are true are exactly those in $R$.
\end{restatable}

We provide a proof for Theorem~\ref{theorem:sound-complete} in an extended version of the paper \cite{drexler-et-al-zenodo2022b}.
The notation $\Ps^*$ is used to denote the \emph{closure} of the problems in $\Ps$; i.e. for any problem $P$ in $\Ps$,
$\Ps^*$ also includes the problems $P'$ that are like $P$ but with initial state $s$ being a solvable state
reachable from the initial state of $P$.

%

\section{Learning Sketches: ASP Implementation}

\begin{listing*}[tb]
    \begin{lstlisting}[escapechar=|, basicstyle=\small\ttfamily]
% C1: construct rules and select features.
{ select(F) } :- feature(F).           { rule(1..max_sketch_rules) }.
{ c_eq(R, F); c_gt(R, F); c_unk(R, F) } = 1 :- rule(R), numerical(F).
{ c_pos(R, F); c_neg(R, F); c_unk(R, F) } = 1 :- rule(R), boolean(F).
{ e_dec(R, F); e_inc(R, F); e_unk(R, F); e_bot(R, F) } = 1 :- rule(R), numerical(F).
{ e_pos(R, F); e_neg(R, F); e_unk(R, F); e_bot(R, F) } = 1 :- rule(R), boolean(F).
% C4 and C7: good and bad state pairs must comply with rules and selected features.
{ good(R, I, S, S') } :- rule(R), s_distance(I, S, S', _).
c_satisfied(R, F, I, S) :- { c_eq(R, F) : V = 0; c_gt(R, F) : V > 0; c_pos(R, F) : V = 1; |\newline|c_neg(R, F) : V = 0; c_unk(R, F) } = 1, rule(R), feature_valuation(F, I, S, V), s_distance(I, S, S', _).
e_satisfied(R, F, I, S, S') :- { e_dec(R, F) : V > V'; e_inc(R, F) : V < V'; |\newline|e_pos(R, F) : V' = 1; e_neg(R, F) : V' = 0; e_bot(R, F) : V = V'; e_unk(R, F) } = 1, rule(R), feature_valuation(F, I, S, V), feature_valuation(F, I, S', V'), |\newline|s_distance(I, S, S', _).
:- { not c_satisfied(R, F, I, S); not e_satisfied(R, F, I, S, S') } != 0, select(F), good(R, I, S, S').
:- { not c_satisfied(R, F, I, S) : select(F); not e_satisfied(R, F, I, S, S') : select(F) } = 0, rule(R), s_distance(I, S, S', _), not good(R, I, S, S').
% C2: there must be one subgoal tuple for each state with unbounded width.
{ subgoal(I, S, T) : tuple(I, S, T) } = 1 :- solvable(I, S), exceed(I, S).
% C3: states underlying subgoal tuples must be good.
:- { good(R, I, S, S') : rule(R) } = 0, subgoal(I, S, T), contain(I, S, T, S').
% C5: good pairs to dead-end states must be at larger distance than subgoal tuple.
:- D <= D', s_distance(I, S, S', D), t_distance(I, S, T, D'), subgoal(I, S, T), |\newline|good(_, I, S, S'), solvable(I, S), unsolvable(I, S').
% C8: ensure acyclicity.
order(I, S, S') :- solvable(I, S), solvable(I, S'), good(_, I, S, S'), order(I, S').
order(I, S) :- solvable(I, S), order(I, S, S') : good(_, I, S, S'), solvable(I, S), solvable(I, S').
:- solvable(I, S), not order(I, S).
% Optimization objective: smallest number of rules plus the sum of feature complexities.
#minimize { C,complexity(F, C) : complexity(F, C), select(F) }.
#minimize { 1,rule(R) : rule(R) }.
    \end{lstlisting}
    \caption{Full ASP code for learning sketches: constraints C1-C8 satisfied.
    Optimization in lines 24-25 for finding a simplest solutions measured
    by number of sketch rules plus sum of feature complexities.}
    \label{listing:asp}
\end{listing*}

We implemented the theory $\sattheory$ expressed by constraints C1--C8 for
learning sketches as an answer set program
\cite{brewka-et-al-cacm2011,lifschitz-2019} in Clingo
\cite{gebser-et-al-2012}. Listing~\ref{listing:asp} shows the code.
We include two approximations for scalability. First, we replace
constraint C8 about termination by a suitable acyclicity condition
\cite{gebser-et-al-kr2014}, that prevents a sequence of states $s_{i+1} \in G^k_R(s_i)$
from forming a cycle, where $R$ is the learned sketch.
Second, we omit constraint C6 that ensures that the width is bounded by
$k$. The choice of the subgoals $t$ still ensures that such subproblems
will have an s-width bounded by $k$. The two approximations are not
critical as even the exact theory $\sattheory$ does not guarantee
acyclicity and sketch width bounded by $k$ on the \emph{test problems}.
However, we will show in our experiments that the learned sketches are
acyclic and have width bounded by $k$ over all instances.
The optimization criterion minimizes the sum of the number of sketch rules
plus the sum of complexities of the selected features.

\newcommand{\mypar}[1]{\paragraph{#1}}

\section{Experiments}

\newcommand{\numtasks}[1]{\small{(#1)}}
\newcommand{\asep}{\hskip 15pt}
\setlength{\tabcolsep}{4pt}

\begin{table*}[tbh]
    \setlength{\cmidrulekern}{15pt}
    \centering
    \resizebox{0.95\textwidth}{!}{
        \begin{tabular}{@{}l@{\asep}rrrrrrrr@{\asep}rrrrrrrr@{\asep}rrrrrrrr@{}}
            & \multicolumn{8}{c}{$w = 0$} & \multicolumn{8}{c}{$w = 1$} & \multicolumn{8}{c}{$w = 2$} \\
            \cmidrule(r){2-9}
            \cmidrule(r){10-17}
            \cmidrule(){18-25}
            Domain & M & T & $|P_i|$ & $|S|$ & $|\cal F|$ & C & $|\Phi|$ & $|R|$ & M & T & $|P_i|$ & $|S|$ & $|\cal F|$ & C & $|\Phi|$ & $|R|$ & M & T & $|P_i|$ & $|S|$ & $|\cal F|$ & C & $|\Phi|$ & $|R|$ \\
        \midrule
        Blocks-clear & 1 & 3 & 1 & 22 & 233 & 2 & 2 & 2 & 1 & 4 & 1 & 22 & 233 & 4 & 1 & 1 & 1 & 3 & 1 & 22 & 233 & 4 & 1 & 1 \\
        Blocks-on & 26 & 14k & 1 & 22 & 1011 & 7 & 3 & 3 & 9 & 105 & 1 & 22 & 1011 & 4 & 2 & 2 & 13 & 146 & 1 & 22 & 1011 & 4 & 1 & 1 \\
        Childsnack & -- & -- & -- & -- & -- & -- & -- & -- & 122 & 228k & 3 & 792 & 629 & 6 & 4 & 5 & -- & -- & -- & -- & -- & -- & -- & -- \\
        Delivery & -- & -- & -- & -- & -- & -- & -- & -- & 17 & 521 & 1 & 96 & 474 & 4 & 2 & 2 & 3 & 18 & 1 & 20 & 287 & 4 & 1 & 1 \\
        Gripper & 2 & 19 & 1 & 28 & 301 & 4 & 2 & 3 & 3 & 60 & 1 & 28 & 301 & 4 & 2 & 2 & 7 & 48 & 1 & 28 & 301 & 4 & 1 & 1 \\
        Miconic & -- & -- & -- & -- & -- & -- & -- & -- & 1 & 5 & 1 & 32 & 119 & 2 & 2 & 2 & 2 & 6 & 1 & 32 & 119 & 2 & 1 & 1 \\
        Reward & 3 & 46 & 2 & 26 & 916* & 6 & 2 & 2 & 1 & 4 & 1 & 12 & 210 & 2 & 1 & 1 & 10 & 95 & 1 & 48 & 427 & 2 & 1 & 1 \\
        Spanner & 12 & 2k & 3 & 227 & 658 & 7 & 2 & 2 & 3 & 22 & 1 & 74 & 424 & 5 & 1 & 1 & 6 & 38 & 1 & 74 & 424 & 5 & 1 & 1 \\
        Visitall & 3 & 54 & 2 & 36 & 722* & 5 & 2 & 2 & 1 & 1 & 1 & 3 & 10 & 2 & 1 & 1 & 1 & 1 & 1 & 3 & 10 & 2 & 1 & 1 \\
        \midrule
    \end{tabular}}
    \caption{Learning step. We show
    the peak memory in GiB after learning (M),
    the time in seconds for solving the ASP in parallel on 32 CPU cores (T),
    the number of training instances used in the encoding ($|P_i|$),
    the total number of states considered in the encoding ($|S|$),
    the number of Boolean and numerical features ($|\cal F|$) where * denotes that the distance feature was included,
    the largest complexity of a feature $f\in\Phi$ ($C$),
    the number of features ($|\Phi|$), and
    the number of sketch rules ($|R|$).
    We use ``--'' to indicate that the learning procedure failed because of insufficient resources.}
    \label{table:learning}
\end{table*}

\begin{table*}[tbh]
    \setlength{\cmidrulekern}{15pt}
    \centering
    \resizebox{0.9\textwidth}{!}{
    \begin{tabular}{@{}l@{\asep}rrrr@{\asep}rrrr@{\asep}rrrr@{\asep}rr@{\asep}rr@{}}
        & \multicolumn{4}{c}{$w = 0$~~~~~} & \multicolumn{4}{c}{$w = 1$~~~~~} & \multicolumn{4}{c}{$w = 2$~~~~~} & \multicolumn{2}{c}{\lama~~~~~~} & \multicolumn{2}{c}{\bfws} \\
        \cmidrule(r){2-5}
        \cmidrule(r){6-9}
        \cmidrule(r){10-13}
        \cmidrule(r){14-15}
        \cmidrule(){16-17}
        Domain & S & T & AW & MW & S & T & AW & MW & S & T & AW & MW & S & T & S & T  \\
        \midrule
        Blocks-clear \numtasks{30} & 30 & 3 & 0.00 & 0 & 30 & 5 & 0.80 & 1 & 30 & 4 & 0.80 & 1 & 30 & 4 & 30 & 6 \\
        Blocks-on \numtasks{30} & 30 & 3 & 0.00 & 0 & 30 & 6 & 1.00 & 1 & 30 & 3 & 0.98 & 1 & 30 & 4 & 30 & 25 \\
        Childsnack \numtasks{30} & -- & -- & -- & -- & 30 & 1 & 0.10 & 1 & -- & -- & -- & -- & 9 & 2 & 5 & 658 \\
        Delivery \numtasks{30} & -- & -- & -- & -- & 30 & 1 & 1.00 & 1 & 30 & 4 & 1.66 & 2 & 30 & 1 & 30 & 1 \\
        Gripper \numtasks{30} & 30 & 4 & 0.00 & 0 & 30 & 3 & 0.50 & 1 & 30 & 656 & 2.00 & 2 & 30 & 1 & 30 & 6 \\
        Miconic \numtasks{30} & -- & -- & -- & -- & 30 & 5 & 0.53 & 1 & 30 & 132 & 2.00 & 2 & 30 & 7 & 30 & 25 \\
        Reward \numtasks{30} & 30 & 4 & 0.00 & 0 & 30 & 2 & 1.00 & 1 & 30 & 1 & 1.00 & 1 & 30 & 2 & 30 & 1 \\
        Spanner \numtasks{30} & 30 & 3 & 0.00 & 0 & 30 & 4 & 0.24 & 1 & 30 & 3 & 0.24 & 1 & 0 & -- & 0 & -- \\
        Visitall \numtasks{30} & 26 & 1360 & 0.00 & 0 & 30 & 20 & 0.00 & 1 & 30 & 21 & 0.00 & 1 & 29 & 213 & 25 & 833 \\
        \midrule
    \end{tabular}}
    \caption{Testing step. We show the number of solved instances (S),
    the maximum time for solving an instance for which all algorithms find a solution, excluding algorithms that solve no instance at all (T),
    the average effective width (AW), and the maximum effective width (MW).}
    \label{table:testing}
\end{table*}

We use the ASP implementation above to learn sketches for several
tractable classical planning domains from the International Planning
Competition (IPC). To learn a sketch, we use two Intel Xeon Gold $6130$
CPUs, holding a total of $32$ cores and $384$ GiB of memory
and set a time limit of seven days (wall-clock time).
For evaluating the learned sketches, we limit time and memory by $30$ minutes and $8$ GiB.
Our source code, benchmarks, and experimental data are available online \cite{drexler-et-al-zenodo2022b}.

\mypar{Data and Feature Generation.} For each domain, we use a PDDL
generator \cite{seipp-et-al-zenodo2022}
to generate a set of training instances small enough to be fully explored
using breadth-first search. For domains with a randomized instance
generator, we generate up to $200$ instances for the same parameter
configuration to capture sufficient variation. Afterwards, we remove
unsolvable instances or instances with more than $10\,000$ states. We
construct the feature pool $\cal F$ by automatically composing description logics
constructors up to a feature complexity of $8$,
using the DLplan library \cite{drexler-et-al-zenodo2022}
and the same bound as the one used by \inlinecite{frances-et-al-aaai2021}.
Compared to theirs, our grammar is slightly richer.
However, since the ``distance'' features
\cite{bonet-et-al-aaai2019} significantly increase the size of the feature
pool, we only include them for the domains where the ASP is unsolvable
without them (indicated by an asterisk in the $|\cal F|$ column in
Table~\ref{table:learning}).
We provide details about the feature grammar in an extended version of the paper \cite{drexler-et-al-zenodo2022b}.
We set the maximum number of sketch rules $m$ to $6$.

\mypar{Incremental Learning.}
Instead of feeding all instances to the ASP at the same time, we
incrementally add only the smallest instance that the last learned sketch
fails to solve. In detail, we order the training instances by the number
of states in increasing order, breaking ties arbitrarily. The first sketch
is the empty sketch $R^{0} = \emptyset$. Then, in each iteration
$i=1,2,\ldots$, we find the smallest instance $I$ in the ordering which
the sketch $R^{i-1}$ obtained in the previous iteration
results in a subproblem of width not bounded by $k$
or in a cycle $s_1, \ldots, s_n = s_1$ of subgoal states $s_{i+1} \in G_R(s_i)$ with $i=1,\ldots,n-1$.
These tests can be performed because the training instances are small.
If such an $I$ exists, we use it as the only training instance
if it is the largest among all previous training instances, otherwise
we add it and proceed with the next iteration.
If no such $I$ exists, the sketch solves all training instances, and we return the sketch.

\mypar{Learning Results.} Table~\ref{table:learning} shows results for the
learning process. We learn sketches of width $0$ in $6$ out of the $9$
domains, of width $1$ in all $9$ domains, and of width $2$ in $8$ out of
$9$ domains. In all cases, very few and very small training instances are
sufficient for learning the sketches as indicated by the number of
instances and the number of states that are actually used (columns $|P_i|$
and $|S|$ in Table~\ref{table:learning}). The choice of bound $k$ has
a strong influence because both the solution space and the number of
subgoals (tuples) grows with $k$. Also, the number of rules decreases as
the bound $k$ increases. The most complex features overall have a
complexity $7$ (requiring the application of 7 grammar rules), the highest
number of features in a sketch is $4$, and the highest number of sketch
rules is $5$, showing that the learned sketches are very compact. We
describe and analyze some of the learned sketches in the next section.

\mypar{Search Results.} Table \ref{table:testing} shows the results of
\siwR searches \cite{drexler-et-al-kr2021} on large, unseen test instances
using the learned sketches.\footnote{Theorem~\ref{thm:sketch_width}
requires a \siwRk search rather than a \siwR search but this a minor
difference for theoretical reasons.} As a reference point, we include the
results for the domain-independent planners LAMA
\cite{richter-westphal-jair2010} and Dual-BFWS
\cite{lipovetzky-et-al-aaai2017}. For Childsnack, Gripper, Miconic and
Visitall, we use the Autoscale 21.11 instances for testing
\cite{torralba-et-al-icaps2021}. For the other domains, where no Autoscale
instances are available, we generate $30$ instances ourselves, with the
number of objects varying between $10$ and $100$. Table \ref{table:testing} shows that the
learned sketches of width $k=1$ yield solutions to the 30 instances of
each of the domains. Some of these domains, such as Childsnack, Spanner and
Visitall, are not trivial for LAMA and Dual-BFWS, which only manage to
solve $9$, $0$, and $29$ instances, and $5$, $0$, and $25$ instances,
respectively. In all cases, the maximum effective width of the \siwR
search is bounded by the width parameter $k$, showing that the properties
of the learned sketches generalize beyond the training set.

\paragraph{Plans.} Comparing the plans found by \siwR to the ones obtained with
LAMA and Dual-BFWS, we see that the \siwR plans have 1) half the length for
Childsnack, Spanner, and Visitall, 2) roughly the same length for Delivery,
Blocks-on, Miconic, and Reward, and 3) about three times the length for
Blocks-clear and Gripper. In Blocks-clear, the width-$0$ sketch defines that
unstacking from any tower (not just the one with the target block) is good until
the target block is clear. In Gripper, transporting a single ball instead of two
balls at a time adds two extra move actions per ball.

\section{Sketch Analysis}

In this section, we describe and analyze some of the learned sketches and prove
that they all have width bounded by $k$ and are acyclic.
We provide proofs of these claims in an extended version of the paper \cite{drexler-et-al-zenodo2022b}.

\subsection{Gripper}

In Gripper  there are two rooms $a$ and $b$. A robot can move
between $a$ and $b$, and pick up and drop balls. The goal is to move all
balls from room $a$ to $b$.
We analyze the three learned sketches for the bounds $k=0,1,2$.
The learned sketch $\Rphi^2$ for $k=2$ has
features $\Phi = \{g\}$, where $g$ is the number of well-placed balls,
and a single rule $r_1 = \prule{}{\INC{g}}$
saying that increasing the number of well-placed balls is good.
These subproblems have indeed width bounded by $2$.
The learned sketch $\Rphi^1$ for $k=1$ has
features $\Phi = \{g_a, g\}$, where $g_a$ is the number of balls in room $a$
and $g$ is the number of balls that are either in room $a$ or $b$. The rules are
\begin{align*}
    r_1 &= \prule{}{\UNK{g},\DEC{g_a}} \\
    r_2 &= \prule{}{\INC{g}}
\end{align*}
where $r_1$ says that picking up a ball in room $a$ is good, and $r_2$ says that dropping a ball in room $b$ is good.
The learned sketch $\Rphi^0$ for $k=0$ has features $\Phi = \{B, c\}$, where $B$ is true iff the robot is in room $b$ and
$c$ is the number of carried balls, and rules
\begin{align*}
    r_1 &= \prule{\EQ{c}}{\UNK{c},\neg B} \\  
    r_2 &= \prule{B}{\UNK{B},\DEC{c}} \\ 
    r_3 &= \prule{\neg B,\GT{c}}{\UNK{B}} 
\end{align*}
where $r_1$ says that moving to room $a$ or picking up a ball in room $a$ is good when not carrying a ball,
$r_2$ says that dropping a ball in room $b$ is good, and
$r_3$ says that moving to room $b$ while carrying a ball is good.
\begin{restatable}{theorem}{theoremgripper}
  The sketches $\Rphi^k$ for Gripper are acyclic and have width $k$ for $k=0,1,2$.
\end{restatable}




\subsection{Blocks-on}

The Blocks-on domain works like the standard Blocksworld domain, where a
set of blocks can be stacked on top of each other or placed on the table.
In contrast to the standard domain, Blocks-on tasks just require to place a
single specific block on top of another block, a width-2 task.
%
%
The learned sketch  $\Rphi$ for $k=1$ has  features $\Phi = \{H, g\}$, where $H$ is true iff a block is being held
and $g$ is the number of blocks that are at their goal location. The sketch  rules are
\begin{align*}
    r_1 &= \prule{}{\neg H, \INC{g}} \\
    r_2 &= \prule{H}{\neg H, \UNK{g}}
\end{align*}
where $r_1$ says that well-placing the block mentioned in the goal
or unstacking towers on the table is good, and $r_2$ says that not holding a block is good.
Starting in states $s_0$ where $H$ is false, rule $r_1$  looks  for subgoal states $s_{1}$ where $H$ is false
and $g$ increased. Starting in states $s'_0$ where $H$ is true, rule $r_2$
is also active which looks for subgoal states $s'_1$ where $H$ is false and $g$ has any value.
From such states $s'_1$, however, rule $r_1$ takes over with $s_0=s'_1$,
and \emph{no subgoal state} where $H$ is true is reached again.
If such states need to be traversed in the solution of the subproblems,
they will not be encountered as subgoal states $s_{i+1}\in G_R(s_i)$.

\begin{restatable}{theorem}{theoremblocks}
    The sketch $\Rphi$ for Blocks-on is acyclic and has width~$1$.
\end{restatable}


\subsection{Childsnack}

In Childsnack \cite{vallati-et-al-ker2018},
there is a kitchen, a set of tables, a set of plates, a set of children,
all sitting at some table waiting to be served a sandwich.
Some children are gluten allergic and hence they must be served
a gluten-free sandwich that can be made with gluten-free bread
and gluten-free content. The sandwiches are produced in the kitchen
but can be moved to the tables using one of several plates.
%
The learned sketch  $\Rphi$ for $k=1$ has
features $\Phi = \{\mathit{sk}, \mathit{ua}, \mathit{gfs}, s \}$, where
$\mathit{sk}$ is the number of \emph{sandwiches} at the \emph{kitchen},
$\mathit{ua}$ is the number of \emph{unserved} and \emph{allergic} children,
$\mathit{gfs}$ is the number of \emph{gluten-free sandwiches}, and
$\mathit{s}$ is the number of \emph{served} children.
The sketch rules are
\begin{align*}
    r_1 &= \prule{}{\UNK{\mathit{sk}},\UNK{\mathit{ua}},\INC{\mathit{gfs}},\UNK{s}} \\
    r_2 &= \prule{}{\DEC{\mathit{sk}},\UNK{\mathit{ua}},\UNK{\mathit{gfs}},\UNK{s}} \\
    r_3 &= \prule{}{\UNK{\mathit{sk}},\DEC{\mathit{ua}},\UNK{\mathit{gfs}},\UNK{s}} \\
    r_4 &= \prule{\EQ{\mathit{ua}}}{\INC{\mathit{sk}},\UNK{\mathit{ua}},\UNK{\mathit{gfs}},\UNK{s}} \\
    r_5 &= \prule{\EQ{\mathit{ua}}}{\UNK{\mathit{sk}},\UNK{\mathit{ua}},\UNK{\mathit{gfs}},\INC{s}}
\end{align*}
which say that making a gluten free sandwich is good,
moving a sandwich from the kitchen on a tray is good,
serving a gluten-allergic child is good,
making any sandwich is good if all gluten-allergic children have been served,
serving any child is good if all gluten-allergic children have been served.
\begin{restatable}{theorem}{theoremchildsnack}
    The sketch $\Rphi$ for Childsnack is acyclic and has width~$1$.
\end{restatable}







\subsection{Miconic}

In Miconic \cite{koehler-schuster-aips2000}, there are
passengers, each waiting at some floor, who want to take an elevator to a
target floor.
The learned sketch $\Rphi$ for $k=1$  has  features $\Phi = \{b,g\}$,
where $b$ is the number of boarded passengers and $g$ is the number of
served passengers. The sketch  rules are
\begin{align*}
    r_1 &= \prule{}{\UNK{b},\INC{g}} \\
    r_2 &= \prule{}{\INC{b},\UNK{g}}
\end{align*}
where $r_1$ says that moving a passenger to their target floor is good,
and $r_2$ says that letting some passenger board is good.
\begin{restatable}{theorem}{theoremmiconic}
    Sketch $\Rphi$ for Miconic is acyclic and has width~$1$.
\end{restatable}



\section{Related Work}

Sketches provide a language for expressing control knowledge by hand or for learning it.
Other  languages have been  developed for the first purpose,  including Golog, LTL, and HTNs;
we focus on  the \textbf{learning problem}.

\medskip

\noindent\textbf{Features, General Policies, Heuristics}.
Sketches were introduced by \inlinecite{bonet-geffner-aaai2021}
and used by hand by \inlinecite{drexler-et-al-kr2021}.
The sketch language is the language of
general policies \cite{bonet-geffner-ijcai2018}
that has been used for learning as well
\cite{martin-geffner-ai2004,bonet-et-al-aaai2019,frances-et-al-aaai2021}.
The description logic features have also been used
to learn linear value functions that can be
used to solve problems greedily
\cite{frances-et-al-ijcai2019,degraaff-et-al-icaps2021wskeps}
and dead-end classifiers \cite{stahlberg-et-al-ijcai2021}.
The use of numerical features that can be incremented
and decremented qualitatively is inspired by QNPs \cite{srivastava-et-al-aaai2011,bonet-geffner-jair2020}.
Other works aimed at learning generalized policies or plans
include planning programs \cite{segovia-et-al-icaps2016},
logical programs \cite{silver-et-al-aaai2020},
and deep learning approaches
\cite{groshev-et-al-icaps2018,bajpai-et-al-neurips2018,toyer-et-al-jair2020},
some of which have been used to learn heuristics
\cite{shen-et-al-icaps2020,karia-srivastava-aaai2021}.

\medskip

\noindent \textbf{HTNs}.
Hierarchical task networks explicitly  decompose  tasks into simpler tasks,
and a number of methods for learning them have been studied \cite{zhuo-et-al-aij2014,hogg-et-al-compint2016}.
These methods, however,  learn decompositions using slightly different inputs
like annotated traces and decompositions.
\inlinecite{jonsson-jair2009} developed a powerful approach for learning
policies in terms of a hierarchy of macros, but the approach is
restricted to domains with certain causal graphs.

\medskip

\noindent \textbf{Intrinsic Rewards and (Hierarchical) RL}.
Intrinsic rewards have been introduced for improving
exploration in  reinforcement learning \cite{singh-et-al-ieeetamd2010},
and several authors have addressed the problem of learning
intrinsic rewards over families of problems.
Interestingly, the title of one of the  papers
is a question ``What can learned intrinsic rewards capture?''
\cite{zheng-et-al-icml2020}.
The answer to this question in our setting is clean and simple:
intrinsic rewards are supposed to capture common subgoal structure.
Lacking a language to talk about families of problems
and a language to talk about subgoal structure, however,
the answer that the authors provide is less crisp:
learned intrinsic rewards are supposed to speed up (deep) RL.
The problem of subgoal structure also surfaces in
\emph{hierarchical RL} that aims at learning
 and exploiting hierarchical structure in RL  \cite{barto-et-al-deds2003,kulkarni-et-al-nips2016}.

\section{Conclusions}

We have developed  a formulation  for learning sketches automatically
given instances of the target class of problems $\Q$ and a bound $k$
on the sketch width. The work builds on prior works that
introduced the ideas of general policies, sketches, problem width,
and description logic features.  The learning formulation
guarantees a  bounded width  on the training instances
but the experiments show that  this and other properties
generalize to entire families of problems  $\Q$,
some of which are  challenging for current domain-independent planners.
The  properties  ensure that all problems in $\Q$
can be solved in polynomial time by a variant
of the \siw algorithm. This is possibly \emph{the first general method for
learning how to decompose planning problems into subproblems with a polynomial complexity
that is  controlled with a parameter}. Three  limitations of the proposed learning
approach are  1)~it cannot be applied to intractable domains,
2)~it yields  large (grounded) ASP programs that are often difficult to solve,
and 3)~it deals with collections of problems encoded in PDDL-like languages, even if the problem of learning subgoal
structure arises in other settings  such as  RL.
Three goals for the future are to improve scalability,
to deal with non-PDDL domains,  taking advantage of recent
approaches  that  learn such domains  from data,
and to use sketches for learning hierarchical policies.
%

\Omit{
but when this is so,  that all problems in $\Q$ can be solved in poly-time
(exponential in $k$), if a given theory is satisfiable.
We have also provided an ASP program that captures
an approximation  of this formulation

as compact representations of the language of policy sketches.
We cast the learning problem as combinatorial optimization.
We showed analytically and empirically that the
learned sketches generalize well across a whole domain
and solve domains where state-of-the-art planners fail.
The learned knowledge is precise,
making it easier to transfer knowledge from one problem to another.
Furthermore, a suitable sketch solves a whole domain in low polynomial time.
Our long-term goal is to learn policy sketches automatically when given
the same inputs as DRL algorithms, where there is no state representation
language. Recent works that learn first-order symbolic languages from
black-box states or from states represented by images
\cite{asai-icaps2019,asai-muise-ijcai2020,bonet-geffner-ecai2020} are
important first steps in that direction.
}

\section{Acknowledgements}
This work was partially supported by an ERC Advanced Grant (grant
agreement no.\ 885107), by project TAILOR, funded by EU Horizon 2020
(grant agreement no.\ 952215), and by the Wallenberg AI,
Autonomous Systems and Software Program (WASP)
funded by the Knut and Alice Wallenberg Foundation.
Hector Geffner is a Wallenberg Guest Professor at
Link{\"o}ping University, Sweden.
The computations were enabled by resources provided by
the Swedish National Infrastructure for Computing (SNIC),
partially funded by the Swedish Research Council through
grant agreement no.\ 2018-05973.

\section{Appendix}

\appendix

In this section, we describe the feature grammar, and we provide proofs for the theorems
on the soundness and completeness of the theory and the properties of the learned sketches.

\section{Feature grammar}

The feature grammar is constructed by iteratively composing
description logics grammar rules \cite{baader-et-al-2003}
and additional grammar rules for Boolean and numerical features.
The feature complexity is then measured as the number of grammar rules used in the composition.
We first define the description logics grammar rules
and then define the grammar rules for Boolean and numerical features.

\subsection{Concepts and Roles}

In description logics \cite{baader-et-al-2003}, we have concepts and roles where concepts represent unary relations,
and roles represent binary relations over the universe $\Delta$, that is, the set of objects occurring in the instances.
The concepts and roles we use are similar to the ones used in work on computing general policies \cite{frances-et-al-aaai2021}.
However, we compute a richer pool of primitives by projecting $n$-ary predicates to concepts and roles.
Consider a state $s$ in an instance $I$.
For every $n$-ary predicate $p$ in $I$ and $1\leq i\leq n$,
there is a concept that evaluates to the set of objects that occur in
the $i$-th position of the respective ground atoms of $p$ that are true in $s$.
Similarly, for every $n$-ary predicate $p$ in $I$ and $1\leq i<j\leq n$,
there is a role that evaluates to the set of pairs of objects
that occur in the $i$-th and $j$-th position of the respective ground atoms of $p$ that are true in $s$.
Furthermore, for every such primitive concept and role as above,
there is a respective goal version that is evaluated in the goal instead of the state $s$.
Next, for each constant $x$ in a domain, there is a concept that evaluates to the singleton set $\{x\}$.
Last, there are the following additional primitive and compositional concepts and roles.
Consider concepts $C, D$, and roles $R, S$. We have

\begin{itemize}
    \item the universal concept $\top$ with denotation $\top^s\equiv \Delta$,
    \item the bottom concept $\bot$ with denotation $\bot^s\equiv \emptyset$,
    \item the role-value mapping $R = S$ with denotation $(R = S)^s\equiv \{a\in\Delta\mid (a,b)\in R^s\leftrightarrow (a,b)\in S^s\}$,
    \item the concept intersection $C\sqcap D$ with denotation $(C\sqcap D)^s = C^s\cap D^s$,
    \item the concept negation $\neg C$ with denotation $(\neg C)^s\equiv \Delta\setminus C^s$,
    \item the existential abstraction $\exists R.C$ with denotation $(\exists R.C)^s\equiv \{a\in\Delta\mid\exists b : (a,b)\in R^s\land b\in C^s\}$,
    \item the universal abstraction $\forall R.C$ with denotation $(\forall R.C)^s\equiv \{a\in\Delta\mid\forall b : (a,b)\in R^s\rightarrow b\in C^s\}$,
    \item the role inverse $R^{-1}$ with denotation $(R^{-1})^s\equiv \{(b,a)\mid (a,b)\in R^s\}$,
    \item the role restriction $R : C$ with denotation $(R : C)^s\equiv R^s\sqcap (\Delta\times C^s)$,
    \item the role composition $R\circ S$ with denotation $(R\circ S)^s\equiv \{(a,c)\in\Delta\times\Delta\mid (a,b)\in R^s\land (b,c)\in S^s \}$, and
    \item the transitive closure role $R^+$ with denotation $(R^+)^s\equiv\bigcup_{n\geq 1} (R^s)^n$ where the iterated composition is defined as $(R^s)^0 = \{(d,d)\mid d\in C^s\}$ and $(R^s)^{n+1} = (R^s)^n\circ R^s$.
\end{itemize}
We place additional restrictions on the above grammar.
Similar to \inlinecite{frances-et-al-aaai2021}, we do not include role composition,
and allow the transitive closure role, inverse role, and restrict role on primitive concepts and roles.

\subsection{From Concepts and Roles to Features}
The concepts and roles from the grammar above
represent more complex derived predicates.
We are interested in measuring a change
in the number of ground atoms over such derived predicates.
For this purpose, we define the following Boolean and numerical grammar rules
as an additional level of composition.
Consider a state $s$ in an instance $I$.
For every nullary predicate $p$, there is a Boolean feature
that is true in $s$ iff the ground atom of $p$ is true in $s$.
Next, let $C, D$ be concept, $R$ be a role, and $X$ be either a concept or role. We have

\newcommand{\fempty}[1]{\mathit{Empty}(#1)}
\newcommand{\fcount}[1]{\mathit{Count}(#1)}
\newcommand{\fdistance}[3]{\mathit{Distance}(#1, #2, #3)}
\begin{itemize}
    \item Boolean empty feature $\fempty{X}^s$ is true iff $|X^s|=\emptyset$,
    \item numerical counting feature $\fcount{X}^s\equiv |X^s|$, and
    \item numerical distance feature $\fdistance{C}{R}{D}$ is smallest $n\in\mathbb{N}_0$, s.t.,
    there are objects $x_0,\ldots,x_n$ with $x_0\in C^s,x_n\in D^s$, and $(x_{i-1},x_i)\in R^s$ for $i = 1,\ldots,n$.
    If no such $n$ exists then it evaluates to $\infty$.
\end{itemize}
We also place restrictions on this part of the grammar.
Our distances features are richer than those used by \inlinecite{frances-et-al-aaai2021}
as we allow arbitrary concepts $C$ in them,
and $R$ can have at most complexity $2$,
meaning that $R$ can take the restriction as well but also other roles.

\section{Proofs}

{\theoremtheory*}

The notation $\Ps^*$ is used to denote the \emph{closure} of the problems in $\Ps$; i.e. for any problem $P$ in $\Ps$,
$\Ps^*$ also includes the problems $P'$ that are like $P$ but with initial state $s$ being a solvable non-goal state that is reachable
from the initial state of $s$.

\noindent \emph{Proof:} ($\Leftarrow$; soundness). Let $\sigma$ be a truth assignment that satisfies $\sattheory$
and let $R_\sigma$ be the sketch encoded in $\sigma$; i.e., the rule $C_i \mapsto E_i$ has conditions $f=v$ for the
atoms $\satcond{i}{f}{v}$ true in $\sigma$, and effects $f$ (resp. $f$) if $\sateff{i}{f}{v}$ with $f$ Boolean and $v$ is $true$ (resp. $false$),
and effects $\DEC{f}$, $\INC{f}$, or $\UNK{f}$ if $f$ is numerical, and $v$ is $true$, $false$, and `$?$' resp. Since the class of problems
$\Ps$ is closed, in order to prove that the sketch $R_\sigma$ has width $\leq k$ it suffices to show that for any problem $P$ in $\cal P$
with initial state $s$, the problem $P'$ that is like $P$ but with goals $G(P) \cup G_{R_\sigma}(s)$, has width $\leq k$.
For this, some atom $\satsubgoal{s}{t}$ must be true for a tuple $t$ with size and width $\leq k$, and therefore, and all
the states $s'$ that result from optimal plans from $s$ to $t$, will be such that $(s,s')$ satisfies some rule in $R_\sigma$,
and therefore all such states $s'$ will belong to $G(s)$. Moreover, such states $s'$ will be the closest states in $G(s)$ from $s$,
as enforced by constraint C6 above and are not dead-ends as enforced by constraint C5,
and therefore, since the width of $t$ is $\leq k$, then the width of problem $P'$ is also bounded by $k$.

\noindent \emph{Proof:} ($\Rightarrow$; completeness)
One must show that a sketch $R$ yields an assignment $\sigma_R$ that satisfies the theory $\sattheory$.
First, in the assignment $\sigma_R$, the truth of all the atoms $\satselect{f}$,
$\satcond{i}{f}{v}$, $\sateff{i}{f}{v}$, and $\satrule{s}{s'}{i}$
is clearly determined by $R$ such that C1 is satisfied.
Second, since $w_R(\Ps^*)\leq k$, it means that all the problems $P'$ with initial state $s$
and goal states among those of $P$ and $G_R(s)$ have width $\leq k$,
and therefore, that for each such state $s$, there is an admissible
chain $t_0, \ldots, t_m$ of size $\leq k$ leading to such goal states optimally.
It suffices to make the atoms $\satsubgoal{s}{t_m}$ true
along with all atoms $\satsubgoals{s}{t_m}{s'}$ for the states $s'$
that result from optimal plans from $s$ to $t_m$ such that constraints C2 and C3 are satisfied.
Third, for every subgoal $s'\in G_R(s)$ from state $s$ with $s\notin S_G$
where $\satsubgoals{s}{t_m}{s'}$ is satisfied there is at least one compatible rule $i$
because otherwise, $s'$ would not be a subgoal in the first place,
meaning that $\satrule{s}{s'}{i}$ can be set to true, satisfying constraint C4.
Forth, due to termination, we know that the sketch can not define a closest subgoal state
that is unsolvable because once in an unsolvable state, any acyclic state sequence
will never end in a goal state such that constraint C5 is satisfied.
Fifth, since the subgoals $\satsubgoal{s}{t_m}$ are reached optimally that means that there cannot be
some state $s'$ that is closer than the subgoal $t_m$ satisfying $\satrule{s}{s'}{i}$ for some rule $i$.
Therefore, C6 is satisfied because those states $s'$ are at the same distance or further away.
Sixth, according to the semantics of sketches, we set every $\satrule{s}{s'}{i}$ to true
for every rule $i$ iff $(s, s')$ is compatible with rule $i$ such that C7 is satisfied.

\begin{theorem}
    \label{thm:gripper_2}
    The sketch $\Rphi^2$ for Gripper is acyclic and has width $2$.
\end{theorem}
\noindent \emph{Proof:} The sketch is acyclic
because the number of balls in room $b$ is incremented but is never decremented.
Next, we show that the sketch width is $2$.
If the robot is at $b$ without carrying a ball and there is at least one ball at $a$,
then we have the admissible chain
\begin{align*}
    &(\gatrobot{b},(\gatrobot{a},\gcarrying{x}), \\
    &(\gatrobot{b},\gcarrying{x}),(\gatrobot{b},\gat{x}{b}))
\end{align*}
where $r_1$ defines the subgoal $\gat{x}{b}$.
In all other cases, where the robot is already at $a$, carrying a ball at $a$,
or carrying a ball at $b$ can be described with respective suffixes of the above admissible chain.
The largest tuple is $2$, implying that the sketch width is $2$.

\begin{theorem}
    \label{thm:gripper_1}
    The sketch $\Rphi^1$ for Gripper is acyclic and has width $1$.
\end{theorem}
\noindent \emph{Proof:} rule $r_1$ decrements $g_a$ which is never
incremented by some other rule. Hence, we can remove it.
Rule $r_2$ increments $g$ which no remaining rule decrements.
Hence, $g$ increases finitely many times and the sketch is acyclic.
Next, we show that the sketch width is $1$.
If the robot is at $b$ without carrying a ball and there is at least one ball at $a$,
then we have the admissible chain $(\gatrobot{b},\gatrobot{a},\gcarrying{x})$
where $r_1$ defines the subgoal $\gcarrying{x}$.
If the robot is already at $a$, then the suffix $(\gatrobot{a},\gcarrying{x})$ applies.
If the robot is at $a$ and carrying a ball,
then we have the admissible chain $(\gatrobot{a},\gatrobot{b},\gat{x}{b})$
where $r_2$ defines the subgoal $\gat{x}{b}$.
If the robot is already at $b$, then the suffix $(\gatrobot{b},\gat{x}{b})$ applies.
The largest tuple is $1$, implying that the sketch width is $1$.

\begin{theorem}
    \label{thm:gripper_0}
    The sketch $\Rphi^0$ for Gripper is acyclic and has width $0$.
\end{theorem}
\noindent \emph{Proof:}
we begin with showing acyclicity.
We also consider the feature $g$ that counts the number of balls at $b$.
The sketch decreases $g$ but never increases $g$ because
picking balls in room $a$ is a subgoal defined by $r_1$,
dropping balls in room $b$ is a subgoal defined by $r_2$,
moving from $b$ to $a$ if not carrying a ball is subgoal defined by $r_1$,
and moving from $a$ to $b$ if carrying a ball is subgoal defined by $r_3$.
Hence, the sketch is acyclic.
Next, we show that the sketch width is $0$.
If the robot is at $b$ and not carrying a ball but there is a ball at $a$,
then we have the admissible chain $(\gatrobot{b},\gatrobot{a})$
where $r_1$ defines the subgoal $\gatrobot{a}$.
If the robot is at $a$ instead not carrying a ball $x$,
then we have the admissible chain $(\gat{x}{a},\gcarrying{x})$
where $r_1$ defines the subgoal $\gcarrying{x}$.
If the robot is at $a$ and carrying a ball,
then we have the admissible chain $(\gatrobot{a},\gatrobot{b})$
where $r_3$ defines the subgoal $\gatrobot{b}$.
If the robot is at $b$ instead carrying ball $x$,
then we have the admissible chain $(\gcarrying{x},\gat{x}{b})$
where $r_2$ defines the subgoal $\gat{x}{b}$.
The largest tuple is $1$, and the subproblems are solved in a single step implying that the sketch width is $0$.

\theoremgripper*
\noindent \emph{Proof:}
follows from Theorems~\ref{thm:gripper_2},\ref{thm:gripper_1}, and \ref{thm:gripper_0}.

\theoremblocks*
\noindent \emph{Proof:} rule $r_2$ sets $H$ to false which no
other rule sets to true such that we can eliminate $r_2$.
The remaining rule $r_1$ defines the subgoal of either
moving a block on the table or stacking the block mentioned in the goal on top of the correct block.
Both cases can only happen finitely many times because there are finitely many blocks.
Hence, the sketch is acyclic.
Next, we show that the sketch width is $1$.
If some block $b$ is being held, then we have the admissible chain
$(\bholding{b},\bontable{b})$ where $r_2$ defines the subgoal $\bontable{b}$.
For the remaining, we can assume that no block is being held.
If there is block $b$ above $b'$, i.e., $\bon{b}{b'}$ holds,
and this is not mentioned in the goal, then we have the admissible chain
$(\bon{b}{b'},\bholding{b},\bontable{b})$
where $r_1$ defines the subgoal $\bontable{b}$.
If both $b, b'$ are clear and $b$ must be stacked on top of $b'$ in the goal,
then we have the admissible chain $(\bclear{b},\bholding{b},\bon{b}{b'})$
where $r_1$ defines the subgoal $\bon{b}{b'}$.
The largest tuple is $1$, implying that the sketch width is $1$.

\theoremchildsnack*
\noindent \emph{Proof:} we begin with showing acyclicity.
Rule $r_1$ can be used as many times as there are ingredients to make
gluten-free sandwiches because the available ingredients decrease.
Similarly, $r_4$ can be used as many times
as there are ingredients to make sandwiches.
Since the number of sandwiches in the kitchen is finite,
and sandwiches can be put on a tray and served,
but not be moved from tray to kitchen,
$r_2$ can be used finitely many times
to put a sandwich on a tray.
Sandwiches are served first to gluten allergic children using $r_3$
and to other children using $r_5$ at most as many times as there are children.
Hence, the sketch is acyclic.
Next, we show that the width is $1$.
If there are ingredients to make a gluten-free sandwich $s$,
then we have the admissible chain $(\cnotexist{s},\cglutenfreesandwich{s})$
where $r_1$ defines the closest subgoal $\catkitchensandwich{s}$.
If all gluten allergic children are served, and there are ingredients to make a sandwich $s$,
then we have the admissible chain $(\cnotexist{s},\catkitchensandwich{s})$
where $r_4$ defines the closest subgoal $\catkitchensandwich{s}$.

Now assume, that no more sandwich can be made as above,
and there is a sandwich $s$ and a tray $t$ in the $\ckitchen$,
then we have the admissible chain $(\cat{t}{\ckitchen},\contray{s}{t})$
where $r_2$ defines the closest subgoal $\contray{s}{t}$.
If $t$ is instead at a table $x$ and there is
no tray with a sandwich on the same table as a child that can be served with it,
then we have the admissible chain $(\cat{t}{x},\cat{t}{\ckitchen},\contray{s}{t})$
where $r_2$ defines the closest subgoal $\contray{s}{t}$.

Now assume, that no more sandwich can be made as above,
and there is a gluten free sandwich $s$ on tray $t$
at the same table $x$ as an unserved gluten allergic child $c$,
then we have the admissible chain $(\cat{t}{x},\cserved{c})$
where $r_3$ defines the closest subgoal $\cserved{c}$.
If $t$ is instead at a place $x'$ including the kitchen different than $x$
then we have the admissible chain $(\cat{t}{x'},\cat{t}{x},\cserved{c})$
where $r_3$ defines the closest subgoal $\cserved{c}$.
Furthermore, if we drop the requirement for $s$ to be gluten-free,
child $c$ is not gluten allergic and require that
there are no more unserved gluten allergic children,
then the same admissible chains also hold
where $r_5$ defines the closest subgoals.

An unsolvable state is reached, if either a non-gluten-free sandwich
is made using gluten-free ingredients
or if serving gluten-free sandwiches to non-gluten allergic children
such that not all gluten allergic children can be served.
However, the sketch defines making gluten-free sandwiches and serving gluten allergic children first
with rules $r_1,r_2,r_3$ such that no closest subgoal state is unsolvable.
The largest tuple is $1$, implying that the sketch width is $1$.

\theoremmiconic*
\noindent \emph{Proof:} the feature $g$ can only be increased
because if a passenger is served, this remains true forever.
The feature $b$ can only be decreased once for every passenger after unboarding and serving it.
Hence, the sketch is acyclic.
Next, we show that the width is $1$.
We define the distance between floors $f_i$ and $f_j$
as $d(f_i,f_j) = |i-j|$.
If there is an unserved passenger $p$ waiting at floor $f_i$,
the elevator is at floor $f_j$,
and for all boarded passengers $p'$ with target floor $f'$
holds that $d(f_i,f_j)\leq d(f',f_j)$,
then we have the admissible chain $(\mliftat{f_i},\ldots,\mliftat{f_j},\mboarded{p})$
where $r_2$ defines the subgoal $\mboarded{p}$.
If there is a boarded passenger $p$ with target floor $f_j$ in the elevator at floor $f_i$,
and for all unserved and unboarded passengers $p'$ waiting at floor $f'$
holds that $d(f_j,f_i)\leq d(f', f_i)$,
then we have the admissible chain $(\mliftat{f_i},\ldots,\mliftat{f_j},\mserved{p})$
where $r_1$ defines the subgoal $\mserved{p}$.
The largest tuple is $1$, implying that the sketch width is $1$.

\section{Additional Sketches with Proofs}

In this section, we provide details about domains
and learned sketches about some additional domains.

\subsection{Visitall}

In the Visitall domain, there is a rectangular grid of places and a robot
located at some random position in the grid. The objective is to visit
each place at least once.
The learned sketch $\Rphi$ for $k=1$ has features $\Phi = \{v\}$,
where $v$ is the number visited places, and the single rule
\begin{align*}
    r &= \prule{}{\INC{v}}
\end{align*}
\vspace{0.2cm}
which says that visiting another  place is ``good''  (a subgoal).
The feature $v$ represents in this case  a goal counter,
and the sketch, a simple goal serialization.
The following result implies that the sketch generalizes over the whole domain:
\begin{theorem}
    The sketch $\Rphi$ for Visitall is acyclic and has width $1$.
\end{theorem}
\noindent \emph{Proof:} the sketch is acyclic
because the feature that counts the number of visited places $v$ is incremented but never decremented.
Next, we show that the sketch width is $1$.
For every non-goal state, the sketch defines the subgoal of moving
from the current position $c_i$ to the closest cell $c_j$.
In this case, we have the admissible chain $(\vatrobot{c_i},\ldots,\vatrobot{c_j})$
where $r$ defines the subgoal $\vatrobot{c_j}$.  
The largest tuple is $1$, implying that the sketch width is $1$.

\subsection{Spanner}

In the Spanner domain, there is a one-way path of locations between a shed
and a gate, spanners distributed over these locations, a set of loose nuts
at the gate, and a man that is initially in the shed. The man can move
along the path towards the gate but not backwards, pick up spanners, and
tighten a loose nut when he is at the gate and carrying a spanner. Each
spanner can only be used once. The objective is to tighten all nuts.
The learned sketch $\Rphi$ for $k=1$ has features $\Phi = \{a\}$, where
$a$ is the number of unpicked spanners plus the number of loose nuts,
and the single rule
\vspace{0.2cm}
\begin{align*}
    r &= \prule{}{\DEC{a}}
\end{align*}
\vspace{0.2cm}
which encodes that picking up a spanner or tightening a nut is good.

\begin{theorem}
    The sketch $\Rphi$ for Spanner is acyclic and has width $1$.
\end{theorem}
\noindent \emph{Proof:} the sketch is acyclic
because feature $a$ is decremented but never incremented.
Next, we show that the sketch width is $1$.
If the man $m$ is at some location $l_i$ and there is a spanner $p$ at location $l_j$ with $i < j$,
then we have the admissible chain $(\sat{m}{l_i},\ldots,\sat{m}{l_j},\scarrying{p})$
where $r$ defines the subgoal $\scarrying{p}$.
If the man $m$ is at the location $l_i$ and there is no spanner available at some other location
but there is a loose nut at gate $g = l_n$, then we have the admissible chain
$(\sat{m}{l_i},\ldots,\sat{m}{l_n},\stightened{n})$
where $r$ defines the subgoal $\stightened{n}$.
An unsolvable state is reached if leaving too many spanners behind.
However, picking up spanner at the current location is the closest subgoal with a distance of $1$
and moving forward to tighten a nut has at least a distance of $2$.
Hence, all spanners will be picked up before moving forward.
The largest tuple is $1$, implying that the sketch width is $1$.

{\bibliography{bib/abbrv-short,bib/literatur,control,extra,bib/crossref-short}}

\end{document}